\documentclass{ecai}
\usepackage{graphicx}
\usepackage[namelimits]{amsmath}
\usepackage{latexsym}
\usepackage{amssymb}
\begin{document}
\title{Pixel-Semantic Revising of Position: One-Stage Object Detector with Shared Encoder-Decoder}
\titlerunning{Pixel-Semantic Revising of Position}
\author{Qian Li\inst{1,2}\orcidID{0000-0001-7454-5001} \and
Nan Guo\inst{1}\orcidID{0000-0002-2250-5224}\and
Xiaochun Ye\inst{1,2}\orcidID{0000-0003-4598-1685} \and
Dongrui Fan\inst{1,2} \and
Zhimin Tang\inst{1,2}
}
\authorrunning{Q. Li et al.}
\tocauthor{Qian~Li, Nan~Guo, Xiaochun~Ye, Dongrui~Fan, and Zhimin~Tang}
\institute{State Key Laboratory of Computer Architecture, Institute of Computing Technology,
Chinese Academy of Sciences, Beijing, China\\
\email{\{liqian18s,guonan,yexiaochun,fandr,tang\}@ict.ac.cn}\\
\and
University of Chinese Academy of Sciences, Beijing, China\\
\email{\{liqian18s,yexiaochun,fandr,tang\}@ict.ac.cn}}
\maketitle \setcounter{footnote}{0}              % typeset the header of the contribution

\begin{abstract}
Recently, many methods have been proposed for object detection. They cannot detect objects by semantic features, adaptively. In this work, according to channel and spatial attention mechanisms, we mainly analyze that different methods detect objects adaptively. Some state-of-the-art detectors combine different feature pyramids with many mechanisms to enhance multi-level semantic information. However, they require more cost. This work addresses that by an anchor-free detector with shared encoder-decoder with attention mechanism, extracting shared features. We consider features of different levels from backbone (e.g., ResNet-50) as the basis features. Then, we feed the features into a simple module, followed by a detector header to detect objects. Meantime, we use the semantic features to revise geometric locations, and the detector is a pixel-semantic revising of position. More importantly, this work analyzes the impact of different pooling strategies (e.g., mean, maximum or minimum) on multi-scale objects, and finds the minimum pooling improve detection performance on small objects better. Compared with state-of-the-art MNC based on ResNet-101 for the standard MSCOCO 2014 baseline, our method improves detection AP of 3.8\%.

\end{abstract}
\section{Introduction}
\begin{figure}[t]
\centering
\includegraphics[width=1.0\columnwidth]{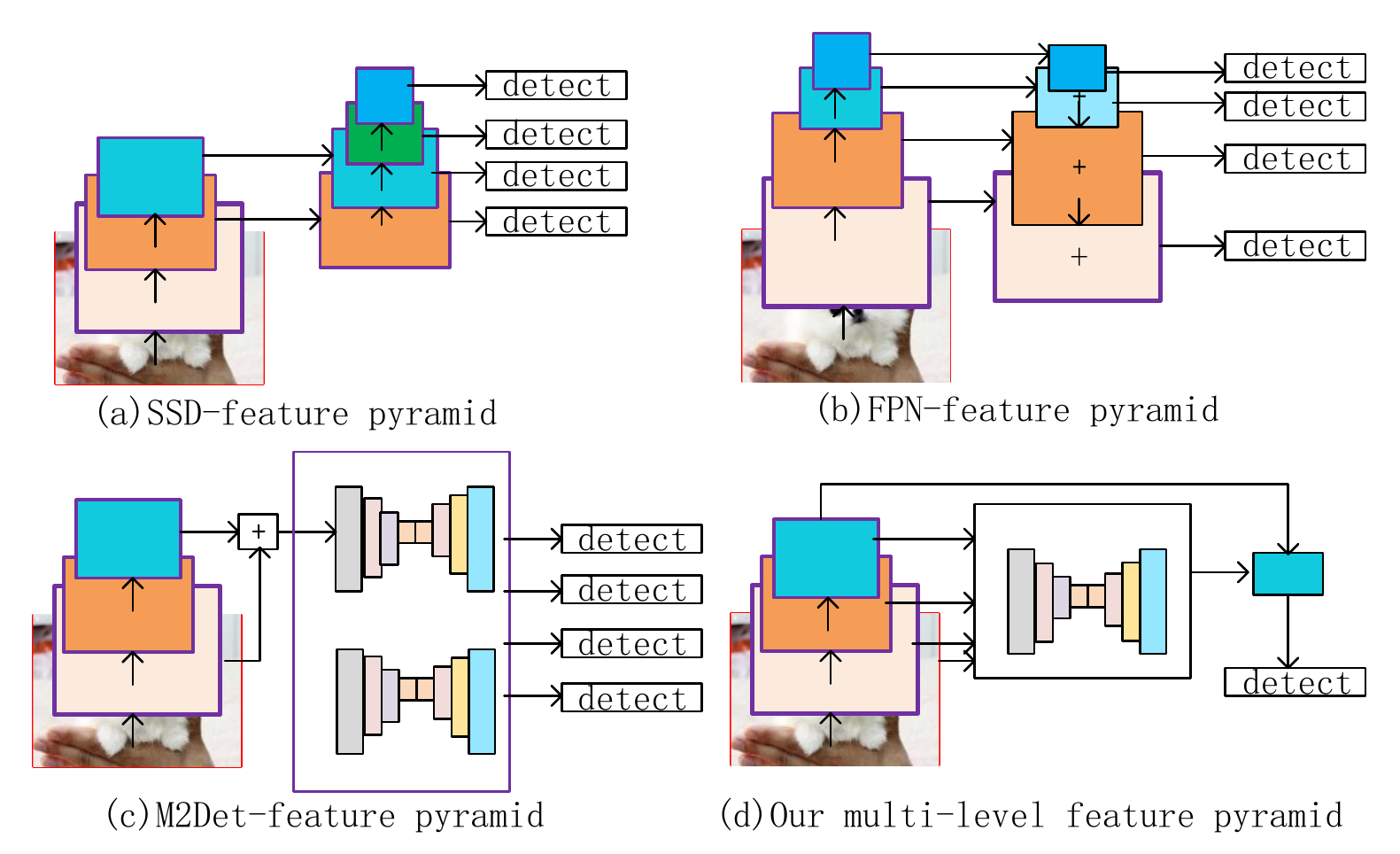} % Reduce the figure size so that it is slightly narrower than the column. Don't use precise values for figure width.This setup will avoid overfull boxes. 
\caption{Illustration of the four feature pyramids, (a) illustrates the feature-based pyramid method \cite{Liu2016SSD} based on anchor for multi-scale objects detection, (b) fuses different horizontal features from top-to-bottom and bottom-to-top to detect multi-scale objects, (c) shows that M2Det  \cite{Zhao2018M2Det} extracts features through many U-shape modules, then combines attention mechanisms to improve detection performance. However, these methods require more time and space. (d) illustrates our multi-scale objects detection with a shared encoder-decoder module for learning shared features on multi-scale objects.}
\label{figure1}
\end{figure}
\begin{figure*}[t]
\centering
\includegraphics[width=.90\textwidth]{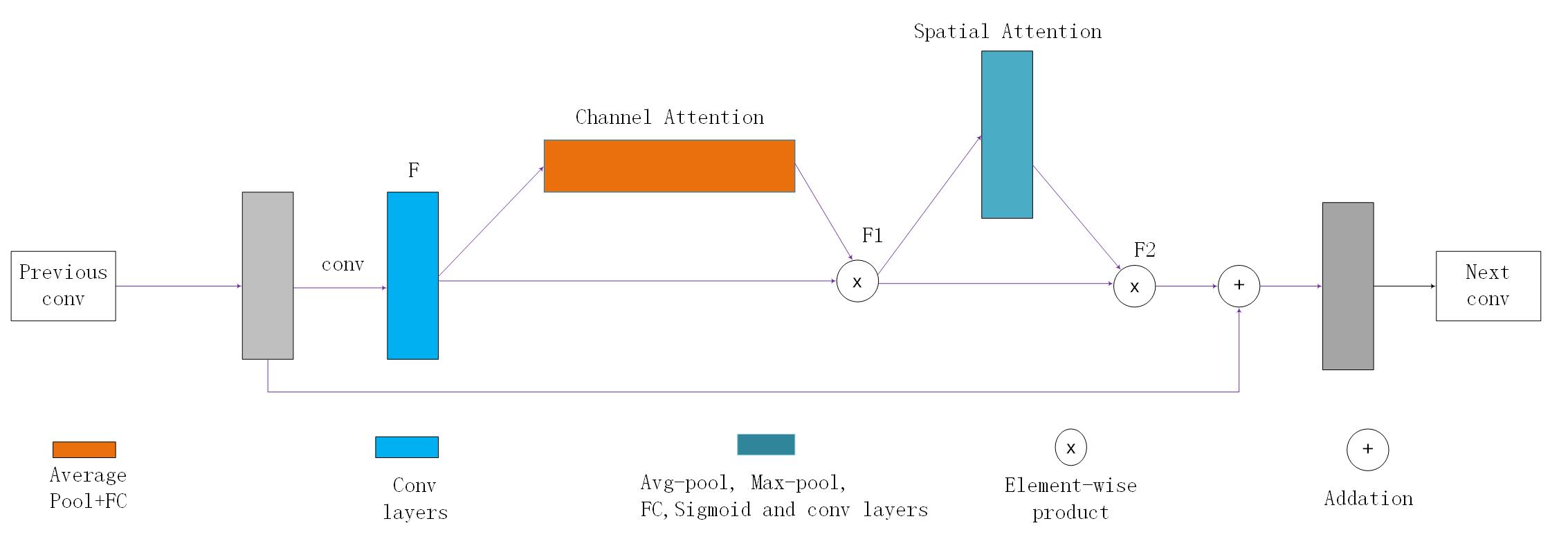} % Reduce the figure size so that it is slightly narrower than the column.
\caption{ResNet\cite{Fei2017Residual} has utilized CBAM. The spatial attention mechanism exploits average-pooling and max-pooling, and followed by a sigmoid layer to normalize features, the channel attention only exploits average-pooling and a sigmoid layer.}
\label{figure4}
\end{figure*}
In recent years, according to the rich representation, CNNs have significantly improved performance of many computer vision tasks (classification, detection and segmentation). Top-5 average precision exceeds 90\% on ImageNet. \cite{xie2019self} proposes a simple self-training method, achieving top-1 average precision of 87.4\%. However, detection performance is poor. There are different methods to improve detection performance. According to region proposals, the object detection methods are divided into the two-stage \cite{girshick14CVPR} \cite{Dai2016R} \cite{Girshick2015Fast} \cite{Ren2017Faster} which mainly focus on the region proposals, and the one-stage \cite{Redmon2017YOLO9000} \cite{Liu2016SSD} \cite{fu2017dssd} \cite{kong2017ron}. The two-stage methods perform better than the one-stage because of multi-scale proposals, but the speed is much slower. According to anchor, detectors are divided into anchor-based \cite{Ren2017Faster}  \cite{Lin2016Feature} \cite{Zhao2018M2Det} which they require more information related to objects, such as, the density, size or shape, and the anchor-free detectors \cite{Yu2016UnitBox} \cite{Huang2015DenseBox} \cite{Redmon2015You} \cite{Redmon2018YOLOv3} \cite{Law2018CornerNet} \cite{Lin2017Focal} \cite{Zhu2019Feature} which use the fully convolution, corner points of the object or adaptive selection of different features, reducing the inference time. Some anchor-free detection methods use feature pyramid to improve multi-scale objects detection performance. Based on feature pyramid and anchor-free methods, we construct a shared encoder-decoder to improve detection performance.

However, for object detection, there are many problems, such as lighting, size, overlapping, etc., resulting in the poor performance. Especially for multi-scale objects, \cite{Lin2016Feature} \cite{SinghAn} \cite{Liu2016SSD} \cite{Zhao2018M2Det} exploit feature pyramids with attention mechanisms to improve the performance, but they require more cost. FPN  \cite{Lin2016Feature} changes anchors for different backbones to implement better, obtaining detection AP of 33.9\%. \cite{Lin2016Feature} successes in obtaining detection AP of 35.8\%, which achieves detection AP of 3\% higher than \cite{Ren2017Faster}. \cite{Xiaowei2018SINet} proposes a new context-aware ROI pooling method, achieving AP of 89.6\% on LSVH. \cite{LiPerceptual} applies a GAN for small objects detection. As shown in Figure \ref{figure1}, feature pyramids \cite{Lin2016Feature} \cite{Liu2016SSD} \cite{Zhao2018M2Det} are obtained by the top-to-bottom, bottom--to-up or both, and parameters from different levels are independent. Inspired by them, we assume that a shared module can be implemented for multi-level features, extracting shared features.

The attention mechanism enhances the key information and suppresses the useless. The attention mainly focuses on the spatial attention, the channel attention or both. Through the attention mechanism, \cite{jaderberg2015spatial} transforms the spatial information into another distribution, retaining the key information. \cite{hu2018squeeze} divides the attention into three parts across channel domain: squeeze, excitation and scale. Compared with the standard residual module, \cite{Fei2017Residual} uses the soft attention and the mask mechanism, and combines the current-level information with the previous. As shown in Figure \ref{figure4}, based on \cite{hu2018squeeze}, \cite{Woo2018CBAM} infers the attention map for two independent dimensions (channels and spatial), and multiplies the attention map with the input feature to improve performance. CBAM \cite{Woo2018CBAM} on ImageNet dataset is 1.76\% higher than the basis ResNet-50. Therefore, for different sizes, we can assume different pooling operation have different detection performance? Can the minimum pooling improve the detection performance on small objects? Our contributions are as follows:
\begin{quote}
\begin{itemize}
\item We propose a shared module, learning feature pyramid by the encoder-decoder with attention mechanism for object detection, extracting the multi-level features by shared parameters to improve the performance on multi-scale objects.
\item We propose a semantic-revised method corresponding to geometric location. Based on the semantic features, our detector can detect objects, adaptively, which is more flexible than state-of-the-state methods of just geometric prediction, and our method is more suitable for the actual scene.
\item This work experiments the impact of the maximum, average, and minimum pooling operations for the small and large objects. The method combining a minimum pooling with \cite{Woo2018CBAM} improves the detection performance on small objects.
\item Based on ResNet-50, our experiment achieves detection AP@0.5 of 49.8\% on standard MSCOCO 2014 benchmark.
\end{itemize}
\end{quote}
\section{Related Works}
\textbf{Feature Pyramid.} For multi-scale objects detection, the traditional methods obtain feature pyramids by different algorithms. SSD \cite{Liu2016SSD} directly predicts features from different levels, and solves the multi-scale problem to a certain extent. \cite{Zhao2018M2Det} uses U-shape module to extract high-level features, then, the method combines the extracted features with the next basis features as the input of the next U-shape. The multi-level weights are independent, resulting in more cost and less correlation. \cite{BaeObject} solves the problem by splitting feature into different modules, and learns the relationship between the original features and each sub-module. Because the relationship between the sub-modules is complex, the relationship becomes a challenge. Therefore, we use a shared module to obtain the multi-level shared features.

\textbf{Encoder-Decoder.} The traditional algorithms \cite{Simonyan2014Very} \cite{Fei2017Residual} learn more discriminative features by deeper network, \cite{Fei2017Residual} introduces the residual module. Each residual module contains two paths, a direct path of the input features and a two-to-three convolution operation on the features, fusing the features obtained by the two paths to get the enhanced features. \cite{sutskever2014sequence} proposes an encoder-decoder to learn text sequences. \cite{badrinarayanan2017segnet} uses an encoder-decoder for classification task. Corresponding to the encoder, a decoder has the same spatial size and the number of channels is same. Therefore, we exploit a shared encoder-decoder to improve detection performance.

\textbf{Attention Mechanism.} Generally, deeper network or attention mechanism can improve the performance. \cite{Fei2017Residual} fuses multi-level features to improve the classification performance. \cite{hu2018squeeze} improves performance by the correlation between channels. A light-weight module \cite{Woo2018CBAM} adopts the channel and spatial attention mechanisms, further improving the performance. We analyze the impact of the detection performance of different attention mechanisms (CBAM with the minimum pooling operation and our attention mechanism) on multi-scale objects. We find that combing CBAM with minimum pooling can improve the detection performance on small objects for standard MSCOCO2014 Dataset.
%\begin{figure}[t]

%\includegraphics[width=1.0\columnwidth]{figure5.pdf} % Reduce the figure size so that it is slightly narrower than the column. Don't use precise values for figure width.This setup will avoid overfull boxes. 
%\caption{Illustrations of our attention mechanism module, we use the average pooling and maximum pooling, followed by a fully connected layer to transform the feature space into the one corresponding to the input, respectively, and the two extracted features are merged, finally, multiplied by the original features as the output.}
%\label{figure5}
%\end{figure}
\section{Our Approach}
\begin{figure*}[t]
\centering
\includegraphics[width=1.0\textwidth]{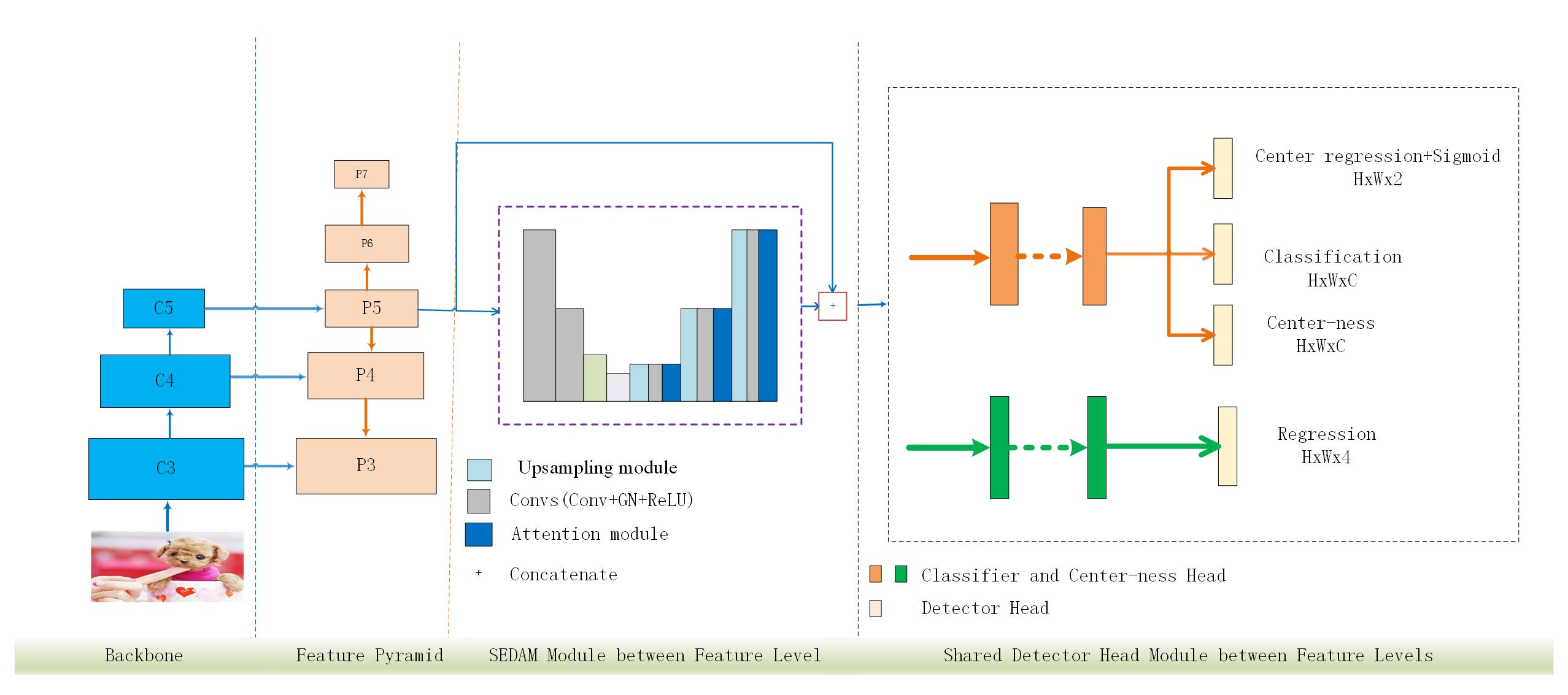}
\caption{An overview of the proposed anchor-free Pixel-Semantic Revising of Position. The architecture exploits the backbone and the shared encoder-decoder module with attention mechanism, obtaining more details for location. Then, the regression prediction produces the four distances (from top boundary to center, center to bottom, right to center, center to left). The semantic-related center prediction branch (center regression +sigmoid in figure) obtains semantic-related center position for revising the pixel-level positions prediction (regression in figure).}
\label{figure3}
\end{figure*}

In this section, as shown in Figure \ref{figure3}, based on ResNet-50, our detection structure includes a shared encoder-decoder module with the attention mechanism for feature pyramid, and a shared detector header with a classification prediction branch, a detection branch, a center-ness branch and a semantic-related center branch which revises regression prediction branch. We introduce a semantic-revised branch to make the detector more suitable for the actual application. For feature pyramids, the shared encoder is down-sampled by a convolution, where the stride is 2, followed by a group normalization and a non-linear activation function. Details as follow.
\subsection{Shared Encoder-Decoder Module with Attention Mechanism (SEDAM)}
As shown in Figure \ref{figure1}, we propose a shared module for multi-level feature pyramids. Since the semantic features within a category are similar, we present that the shared module learns the common features on multi-scale objects for a class, improving the generalization performance. It is a symmetrical structure. In the encoder, features are down-sampled by the convolution with 2 strides and 1 padding, following by a 32 groups normalization and a non-linear activation function. The more the number of layers is, the more the discriminative features extracted are, on the contrary, losing more details about location. In the decoder, the features are up-sampled by a bilinear interpolation, followed by a convolution with 1$\times$1 kernel size and a nonlinear function. Additional, for more useful information, we analyze different attention mechanism in the shared encoder-decoder, including a spatial attention with different pooling operations, a channel attention or the both.

\begin{figure}[t]
\centering
\includegraphics[width=0.9\columnwidth]{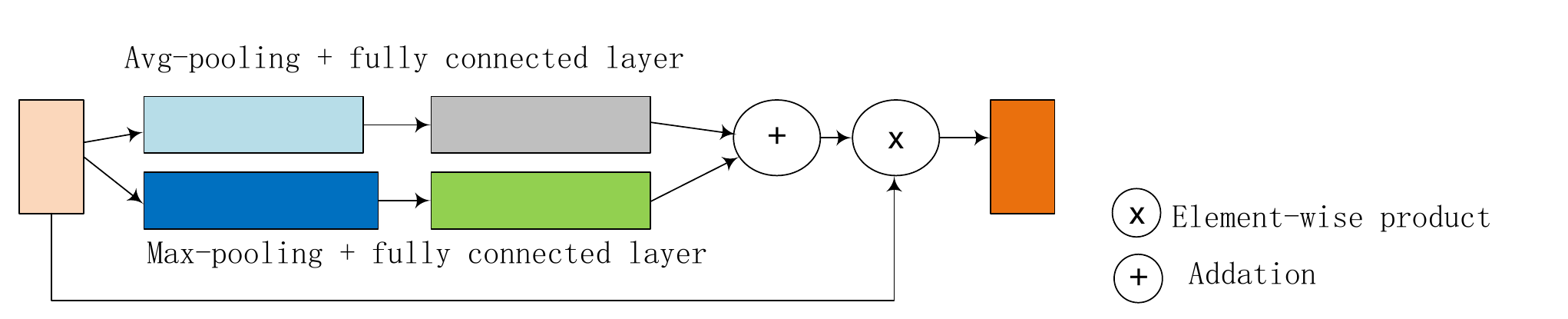} % Reduce the figure size so that it is slightly narrower than the column. Don't use precise values for figure width.This setup will avoid overfull boxes. 
\caption{Illustrations of our attention mechanism module, we use the average pooling and maximum pooling, followed by a fully connected layer, then, multiplied by the original features.}
\label{figure5}
\end{figure}
\textbf{Our Attention Mechanism.} As shown in Figure \ref{figure4}, the state-of-the-art CBAM \cite{Woo2018CBAM} mainly contains channel and spatial attention mechanisms for classification task. We use the channel attention to improve the detection AP, as shown in Figure \ref{figure5}, for our attention mechanism, the features pass two paths (an average-pooling followed by a fully connection layer, and a maximum-pooling followed by a fully connection layer) respectively. We fuse outputs from the two paths, then multiplies with the features which are the input of paths to enhance the key information.
\subsection{Shared Detector Header}
We apply a shared detector header, and regard the fusion of the output of the shared encoder-decoder with the original features as the input to maintain more knowledge about the location and ensure the key information for location. When features of different levels use the same detector header, the detection AP on small objects is better. As shown in Figure. \ref{figure3}, we use semantic-related location prediction to obtain outputs which revise the results from regression branch. We regard geometric position of each bounding box as a $4-D$ vector. The semantic feature can get the semantic-related center location, so that we ensure that the location prediction is related to the semantic information.
\subsection{Margin Regression}
In the feature pyramid, we use four-level features to detect objects. We elaborate the prediction processing of  the $i-$level in detail, other levels are similar. Many candidates of bounding boxes  are obtained at the level $i$.  We define all candidates as {${D_i}$} at the level $i$, where ${D_i}$=(${x^k}_s$, ${y^k}_s$, ${x^k}_m$, ${y^k}_m$, $c_k$) $\in$$R^4\times$\{1,2,3,...,C\}. $C$ is the number of categories, we set it to 80 on MSCOCO, $c_k$ represents the class label in the $k-th$ bounding box. 

We propose a semantic-related location, as shown in the Equation \ref{eq2}. For the semantic center, $B_i$=(${x^j}_i$, ${y^j}_i$) represents the $j-th$ semantic-related center position prediction at the level $i$, the number of the semantic center $B_i$ and the number of the candidates $D_i$ is same. In classification module at the level $i$, if the center proposal position $(x^k_i+x^j_i, y^k_i+y^j_i)$ falls into the truth proposal at the level $i$, the bounding box is a positive example and the class label is  $c_k$. Otherwise, the bounding box is a negative example and the label is 0 (background class).
\begin{equation}
\small
x1= (x^k_i+x^j_i)-x^k_s, y1=(y^k_i+y^j_i)-y^k_s, x2= x^k_m+(x^k_i+x^j_i), y3=y^k_m+(y^k_i+y^j_i)
 \label{eq2}
\end{equation}
Where $(x^k_i, y^k_i)$ denotes the $k-th$ center proposal position at the level $i$. $(x^j_i, y^j_i)$ represents the $j-th$ semantic-related center prediction for revising the $k-th$ center proposal position. $(x^k_s, y^k_s)$ is the left-top margin of the $k-th$ prediction, and  $(x^k_m, y^k_m)$ is the right-bottom margin of the $k-th$ prediction. $(x1, y1)$ and $(x2, y2)$ represent the left-top postion and right-bottom position of the $k-th$ prediction at the level $i$, respectively.
\subsection{Network Configures}
Based on ResNet-50 backbone network, as shown in Figure.\ref{figure3}, the encoder uses three down-sampling modules (a convolution, a group normalization and a ReLU), followed by a smooth layer, the decoder uses three up-sampling modules (a bilinear interpolation, a convolution, a group normalization, a ReLU and the attention mechanism). The channel of the basis features is 256. Therefore, if we set more channels in SEDAM, such as 1024, 4096, etc., the number of channels far exceeds the basis features, resulting in unnecessary computation. In this work, we set it to 640. As shown in Figure \ref{figure3}, we set the input size as 800$\times$800 for backbone.

\textbf{Loss Function.}The structure contains center prediction, regression and classification losses. If a location prediction is closer to the center of the target, the probability value is closer to 1.0. The classification loss is a focal loss with an alpha 0.25 and a gamma 2. Finally, we use a cross-correlation loss with a correlation coefficient to avoid non-overlapping parts. As described in Equation. \ref{eq3}, the loss function details. In our experiment, we set all balance factors ($\gamma$ and $\beta$) to 1.  Details are explained by  \cite{tian2019fcos}. Details as formulate:
\begin{equation}
\begin{split}
L(p_{x,y},d_{x,y})= & \frac{1}{N_{pos}}\sum\limits_{ x,y}L_{cls}(p_{ x,y},{c^\ast}_{ x,y})\\
 &+\frac{\gamma}{N_{pos}}\sum\limits_{ x,y}L_{reg}( d_{x,y},{d^\ast}_{ x,y})\\
&+\frac{\beta}{N_{pos}}\sum\limits_{ x,y}L_{center}(p_{ x,y},{c^\ast}_{ x,y})
\end{split}
\label{eq3}
\end{equation}
Where $L_{cls}(p_{ x,y},{c^\ast}_{ x,y})$ is the cross-entropy classification loss between predicted labels and truth labels. $L_{reg}( d_{x,y},{d^\ast}_{ x,y})$ denotes the regression loss with target center weights between the predicted locations and the target locations, and the weights are related to target margins (the left, the right, the top and the bottom), we regard the IOU-loss as our regression loss. $L_{center}(p_{ x,y},{c^\ast}_{ x,y})$ represents the center loss, which is the cross-entropy loss between center-ness predictions and target center weights. $N_{pos}$ is the number of positive examples which are not background.
%
%\begin{center}
%$L(p_{x,y},d_{x,y})=
%\frac{1}{N_{pos}}\sum\limits_{ x,y}L_{cls}(p_{ x,y},{c^\ast}_{ x,y})$\\
%$+\frac{\alpha}{N_{pos}}\sum\limits_{ x,y}L_{reg}( d_{x,y},{d^\ast}_{ x,y})$\\
%$+\frac{\beta}{N_{pos}}\sum\limits_{ x,y}L_{center}(p_{ x,y},{c^\ast}_{ x,y})$
%\end{center}

\section{Experiments and Results}
In this section, we experiment different detection methods on large-scale standard MSCOCO 2014 benchmark. We experiment 80 classes of train/validation, the training set includes 82783 images, and the validation set includes 40504 images. To compare with the state-of-the-art methods, we compare with traditional methods based on FPN \cite{Lin2016Feature}. In our experiments, we experiment four methods, the A (without a shared encoder-decoder), the B (a shared encoder-decoders with CBAM), the C (a shared encoder-decoder combing CBAM with minimum pooling), and ours (a shared encoder-decoder with our attention mechanism).  

\textbf{Implementation Details.} Based on ResNet-50, our network uses a random gradient descent method for $300k$ iterations, where an initial learning rate, a decay rate and momentum are 0.01, 0.0005, 0.9, respectively. We use ImageNet weights to initialize ResNet-50. For the shared encoder-decoder and detector header, we use a gauss function to initialize weights. When the channel convolution is larger than 32 for the shared encoder-decoder and detector header, we apply group normalization to make the training more stable. In our work, we use 2 TITAN Xp GPUs, 8 batch size for training.
\begin{table*}[t]
\caption{Comparison with using different attention mechanisms, there are four methods, the A (without a shared encoder-decoder), the B (a shared encoder-decoders with CBAM), the C (a shared encoder-decoder combing CBAM with minimum pooling), and ours (a shared encoder-decoder with our attention mechanism).}
\label{table1}
\smallskip
\resizebox{0.98\textwidth}{!}{ % If your table exceeds the column or page width, use this command to reduce it slightly
\centering
\begin{tabular}{|l|l|l|l|l|l|l|l|l|l|l|l|l|l|l|l|l|l|l|l|}
%%%%---------not use single shared encoder decoder with mechansim--
\hline
Method&SED&CBAM&
IOU&
Aera&
person&airplane&bus&train&fire hydrant&stop sign&cat&elephant&bear&zebra&giraffe&toilet&clock\\

\hline
A&-&-&
0.5:0.95&
S&
18.8&23.7&6.66&6.96&20.7&11.0&11.3&21.6&4.9&29.1&24.9&11.0&22.4\\
\hline
B&$\checkmark$&$\checkmark$&
0.5:0.95&S&
19.4&23.2&9.02&7.07&22.5&12.0&10.1&24.0&8.11&28.6&26.2&12.0&22.9\\
\hline
C&$\checkmark$&*&
0.5:0.95&
S&
19.2&25.4&8.76&7.4&20.8&12.1&11.6&23.4&8.17&28.2&25.9&16.7&24.7\\
\hline
\textbf{Ours}&$\checkmark$&-&
0.5:0.95&
S&
%%19.3&24.3&8.79&7.11&21.8&12.2&11.0&23.8&9.38&29.2&25.8&13.0&24.7\\
19.6&23.8&8.26&7.23&21.9&12.3&9.68&23.4&9.41&29.8&26.6&13.6&24.0\\

%%%%------------use single shared encoder decoder with mechansim--
\hline\hline
A&-&-&
0.5:0.95&M&
44.3&40.4&31.9&25.3&52.4&55.6&43.4&44.5&58.9&50.8&54.9&41.7&48.6\\
\hline
B&$\checkmark$&$\checkmark$&
0.5:0.95&
M&
45.2&41.4&34.5&25.5&55.2&56.3&44.5&47.4&62.0&50.6&54.6&44.9&50.2\\
\hline
C&$\checkmark$&*&
0.5:0.95&
M&
45.5&43.7&34.1&28.0&57.4&57.6&43.7&46.4&58.4&51.8&56.0&43.7&49.9\\
\hline
\textbf{Ours}&$\checkmark$&-&
0.5:0.95&
M&
%%45.3&42.7&34.7&24.4&58.0&55.5&44.1&47.1&60.4&51.8&54.8&43.7&49.0\\
45.2&43.1&34.8&23.8&57.5&56.5&43.6&47.3&59.9&51.5&54.8&44.3&48.8\\

\hline\hline
A&-&-&
0.5:0.95&
L&52.5&51.1&63.1&54.3&62.0&77.7&49.4&57.0&59.4&56.4&54.0&49.0&50.5\\
\hline
B&$\checkmark$&$\checkmark$&
0.5:0.95&
L&
55.4&56.4&67.9&57.8&67.8&80.5&55.3&63.3&63.0&58.0&59.8&54.3&53.0\\
\hline
C&$\checkmark$&*&
0.5:0.95&
L&
56.3&58.6&68.2&59.9&69.1&81.4&57.3&63.6&64.6&60.8&60.2&56.7&52.3\\
\hline
\textbf{Ours}&$\checkmark$&-&
0.5:0.95&
L&
%%58.0&60.0&69.8&59.7&68.7&80.6&56.9&64.1&64.7&63.1&63.8&56.7&53.1\\
\textbf{58.4}&\textbf{59.7}&\textbf{69.3}&59.5&\textbf{69.3}&80.6&57.1&\textbf{64.5}&\textbf{64.8}&\textbf{61.7}&\textbf{62.5}&\textbf{57.1}&\textbf{53.0}\\
\hline\hline
A&-&-&
0.5&
-&64.1&70.0&69.1&76.9&74.2&66.8&77.7&76.4&81.8&81.5&80.7&70.3&67.2\\
\hline
B&$\checkmark$&$\checkmark$&
0.5&
-&
68.9&74.0&73.2&80.0&77.5&69.4&81.7&81.2&84.9&83.9&84.8&74.1&69.0\\
\hline
C&$\checkmark$&*&
0.5&
-&
68.5&75.2&72.8&80.7&78.8&69.1&82.3&80.9&84.7&85.1&84.4&76.1&69.1\\
\hline
\textbf{Ours}&$\checkmark$&-&
0.5&
-&
%%69.0&74.1&73.1&78.8&78.4&69.0&81.3&81.3&85.0&84.3&85.3&74.4&68.8\\
69.3&73.5&72.9&79.1&79.0&69.9&81.9&81.0&84.5&83.9&85.0&75.1&68.1\\
\hline\hline
A&-&-&
0.75&-&
33.7&42.6&55.1&54.0&56.2&57.7&52.6&51.9&65.9&52.4&53.6&49.0&36.8\\
\hline
B&$\checkmark$&$\checkmark$&
0.75&
-&
34.5&45.2&59.0&57.2&61.0&59.1&58.1&56.9&72.2&53.2&56.1&53.8&38.4\\

%%%%------------use single shared encoder decoder with mechansim and min--

\hline
C&$\checkmark$&*&
0.75&
-&
35.1&47.7&58.9&59.4&61.2&59.4&59.8&56.3&70.2&53.7&57.8&56.0&39.6\\

%%% our final model based resnet-50 with single channel attention

\hline
\textbf{Ours}&$\checkmark$&-&
0.75&
-&
36.3&47.7&59.4&58.7&61.9&58.7&59.8&57.9&70.9&56.1&58.3&56.9&38.9\\
\hline\hline
A&-&-&
0.5:0.95&-&
35.1&40.5&47.9&48.5&48.6&50.4&47.5&47.9&57.7&49.8&50.5&44.6&37.3\\

\hline
B&$\checkmark$&$\checkmark$&
0.5:0.95&-&
36.9&43.4&52.1&51.6&53.0&52.2&52.4&52.8&61.2&50.7&53.8&49.2&38.7\\

\hline
C&$\checkmark$&*&
0.5:0.95&-&
37.2&45.5&52.2&53.8&54.0&53.0&53.9&52.4&61.7&52.5&54.5&50.6&39.1\\
\hline
\textbf{Ours}&$\checkmark$&-&
0.5:0.95&-&
\textbf{38.0}&45.4&\textbf{52.7}&52.7&54.0&52.4&53.8&\textbf{53.6}&\textbf{62.2}&\textbf{53.0}&\textbf{55.8}&\textbf{51.0}&\textbf{38.7}\\
\hline

\end{tabular}
}
\end{table*}
\subsection{Ablation Studies}
\textbf{The Importance of the Shared Encoder-Decoder.} As mentioned before, the feature pyramid can improve performance on multi-scale objects. As shown in Table \ref{table1}, the method A is poor on small objects. For example, the clock, the stop sign and bear achieve an AP of 22.4\%, 11.0\% and 4.9\%, respectively. We find that ours is better than the A for large and medium objects. For large objects, the person, the airplane, the fire hydrant and the toilet achieve 5.9\%, 8.6\%, 7.3\%, and 8.1\% higher than the A, respectively. As shown in Table \ref{table2}, ours with the semantic-related center is 1.0\% higher on small object detection than the B with the semantic-revised. For small, middle and large objects, ours with semantic-revised module achieves 1.3\%, 1.8\% and 6.3\% higher than the A without semantic-revised module, respectively. Therefore, the shared encoder-decoder with our attention mechanism performances on multi-scale objects better.

\textbf{Comparison of Different Attention Mechanisms.}We think that attention mechanisms can improve the performance for object detection. As shown in Table \ref{table2}, the A with semantic-revised module, the B with semantic-revised module, the C with semantic-revised module and ours with semantic-revised module achieve detection AP@0.5:0.95 of 25.3\%, 27.4\%, 27.8\%, and 28.4\%. The minimum pooling operation is not obvious for improving detection AP. According to different IOU values, the A with semantic-revised module, the B with semantic-revised module, the C with semantic-revised module and ours with semantic-revised module achieve detection AP@0.75 of 24.9\%, 26.7\%, 27.3\%, and 28.1\%, respectively. At the same time, they achieve detection AP@0.5 of 45.4\%, 49.2\%, 49.5\% and 49.8\%. Therefore, the shared encoder-decoder module with our attention mechanism can improve the detection performance. 

As shown in Table \ref{table1}, the C is better than the B on small objects. For clock, toilet and airplane, the C are detection AP of 1.8\%, 4.7\% and 2.2\% higher than the B, respectively. According to Table \ref{table2}, we find that the minimum pooling performs better on small objects  for detection task, and the channel attention mechanism is more suitable to detect multi-scale objects.
%%%%%%%
\begin{table}
\small
\caption{Comparisons of Detection APs(\%) on MS COCO2014 benchmark.}
\label{table2}

\resizebox{0.95\columnwidth}{!}{ % If your table exceeds the column or page width, use this command to reduce it slightly
\smallskip\begin{tabular}{|l|l|l|lll|lll|}

\hline
Method&
Backbone&
Revise&
\multicolumn{3}{l|}{Avg.Precision, IOU:}
&\multicolumn{3}{l|}{Avg.Precision, Area:}
\\

&&&
0.5:0.95&
0.5&
0.75&
S&
M&
L\\

\hline
Faster R-CNN \cite{Ren2017Faster}&
VGG-16&
-&
21.9&42.7&-&-&-&-\\
\hline
OHEM++ \cite{Shrivastava}&
VGG-16&
-&
25.5&45.9&26.1&
7.4&27.7&40.3\\
\hline
SSD \cite{Liu2016SSD}&
VGG-16&
-&
25.1&43.1&25.8&
6.6&25.9&41.4\\
\hline
SSD&MobileNet-v2&-&
22.1&-&-&-&-&-\\
\hline
DSSD321 \cite{fu2017dssd}&ResNet-101&-&
28.0&46.1&29.2&7.4&28.1&47.6\\
\hline
R-FCN \cite{Dai2016R} &ResNet-50&-&
27.0&48.7&26.9&9.8&30.9&40.3\\
\hline
MNC \cite{DaiInstance}&ResNet-101&-&
24.6&44.3&24.8&4.7&25.9&43.6\\
\hline
A&ResNet-50&-&25.1&45.4&24.6&10.5&29.3&32.6\\
\hline
A&ResNet-50&$\checkmark$&25.3&45.4&24.9&10.8&29.2&33.0\\
\hline
B&ResNet-50&-&27.3&49.4&26.5&11.1&30.7&36.8\\
\hline
B&ResNet-50&$\checkmark$&27.4&49.2&26.7&11.5&30.6&36.6\\
\hline
C&ResNet-50&-&27.5&49.5&26.9&11.3&30.9&37.4\\
\hline
C&ResNet-50&$\checkmark$&27.8&49.5&27.3&11.9&31.1&37.3\\
\hline
Ours&ResNet-50&-&28.4&49.9&28.1&11.5&31.2&39.0\\
\hline
Ours&
ResNet-50&$\checkmark$&
\textbf{28.4}&
\textbf{49.8}&28.1&\textbf{11.8}&31.1&38.9\\
\hline

\end{tabular}
}
\end{table}

\textbf{The Importance of the Semantic-Revised.} In this work, we propose a semantic-revised center location. When the network without the semantic-revised center at an inference, the network performs worse on small objects. There are four methods without the semantic-revised, the method A, the method B, the method C, and ours, they are poor detection AP of 0.3\%, 0.4\%, 0.6\% and 0.3\% lower than methods with the semantic-revised center module, respectively. Therefore, we think that the semantic-revised branch makes detection performance better on multi-scale objects, adaptively.
\subsection{Comparison with State-of-the-art Detectors}
To further illustrate that the method which assembles the semantic-revised center branch and the shared encoder-decoder module with attention mechanism can improve detection performance on the multi-scale objects. As shown in Table \ref{table2}, our method is better than \cite{fu2017dssd} and MNC \cite{DaiInstance}, and ours achieve detection AP of 0.4\% and 3.8\% higher than the others, respectively. On the other hand, our method consumes less time and space.  For MSCOCO dataset benchmark, the four methods (the A, the B, the C and ours) are better than traditional detectors \cite{Ren2017Faster} \cite{Shrivastava} \cite{Liu2016SSD} \cite{Dai2016R} \cite{DaiInstance}.
\begin{figure*}[t]
\centering
\includegraphics[width=0.9\textwidth]{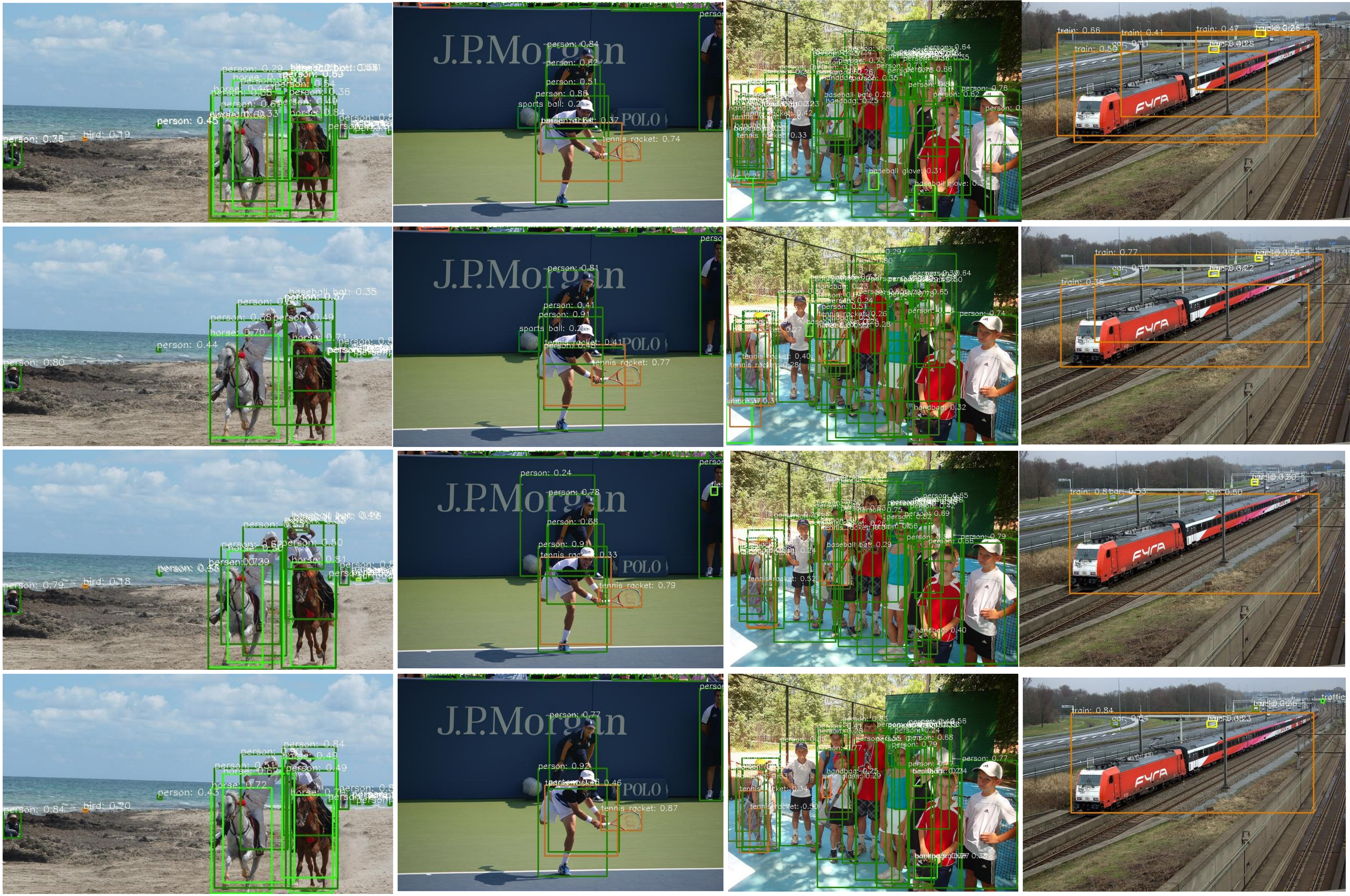} % Reduce the figure size so that it is slightly narrower than the column. Don't use precise values for figure width.This setup will avoid overfull boxes. 
\caption{As shown in the figure, the examples from different methods about a shared encoder-decoder and attention mechanisms, the first two columns show the detection results by the semantic-revised center at an inference, and the last two columns show the ones only by the geometric position. From the first row to the fourth row, these objects detected by the method A, B, C, and ours, respectively.}
\label{figure6}
\end{figure*}
\section{Discussion}
As shown in Table \ref{table1} and Table \ref{table2}, we believe that the attention mechanism plays an important role in objects detection. We find that the minimum pooling performs better on small objects. The minimum pooling extracts much more discriminative features, optimizing the model toward features from small objects, so that the method C performs better than the others on small objects. However, for multi-scale objects, the detection AP of the shared encoder-decoder with channel attention mechanism is higher than others. On the other hand, according to Table \ref{table1}, the shared encoder-decoder can learn the similar semantic features on multi-scale objects. As shown in Figure \ref{figure6}, we use two inference methods, with or without semantic-revised center for all methods (the A, the B, the C and ours). More importantly, our encoder-decoder module can extract the common semantic features on multi-scale objects. However, the semantic distribution between different categories may hurt performance because of the difference of distribution. According to these experiments, we find that our attention mechanism is more effective than the traditional attention mechanism for multi-scale objects detection.
\section{Conclusions}
We propose one-stage anchor-free detector with shared encoder-decoder with attention mechanism, exploiting SEDAM to detect multi-scale objects, adaptively. More importantly, our method uses the semantic-related center to revise the geometric position prediction adaptively, which improves detection performance for the multi-scale objects on the MSCOCO 2014 benchmark, and the semantic-revised branch is more suitable for the actual scene. The attention mechanism with the minimum pooling improves performance on small objects detection better. Therefore, a shared encoder-decoder structure with the attention mechanism can improve detection AP. Our approach reduces cost and performs better than traditional methods for multi-scale objects detection. We believe that our approach can be used to detect multi-scale objects in other basis structures to a certain extent.

\bibliography{ecai-sample-and-instructions}
\bibliographystyle{ecai}
\end{document}